%% file: conference_101719.tex
\documentclass[letterpaper, 10 pt, conference]{ieeeconf}
\IEEEoverridecommandlockouts
\usepackage{cite}
\usepackage{amsmath,amssymb,amsfonts}
\usepackage{algorithmic}
\usepackage{graphicx}
\usepackage{textcomp}
\usepackage{hyperref}
\usepackage{stfloats}
\usepackage{multirow}
\usepackage{bbding}
\usepackage{booktabs}
\usepackage[table,xcdraw]{xcolor}
\usepackage{soul}
\usepackage{tikz}
\usepackage[font=small]{caption}
\usepackage{subcaption}
\usepackage{tabularx}
\usepackage{physics} %

\usepackage{tikz}
\usetikzlibrary{positioning,angles,quotes,shapes,fit,calc,arrows.meta,patterns,patterns.meta,decorations.pathreplacing,calligraphy}

\newcommand{\po}{\phantom{0}}
\usepackage{pgfplots}
\pgfplotsset{compat=1.18,
	every axis/.append style={
		scale only axis,
		enlargelimits=false,
		ylabel near ticks,
		xlabel near ticks,
		title style={yshift=-8pt, font=\footnotesize},
		label style={font=\footnotesize},
		ylabel style={yshift=-3.5pt},
		xlabel style={yshift=2pt},
		tick label style={font=\tiny},
        yticklabel style={xshift=1.5pt},
        }
    }

\usepackage{overpic}

\definecolor{sb_blue_025}{RGB}{199, 221, 236}
\definecolor{sb_orange_025}{RGB}{255, 223, 195}
\definecolor{sb_blue}{RGB}{31, 119, 180}
\definecolor{sb_orange}{RGB}{255, 127, 14}
\definecolor{sb_green}{RGB}{44, 160, 44}
\definecolor{sb_red}{RGB}{214, 39, 40}
\definecolor{sb_purple}{RGB}{148, 103, 189}
\definecolor{sb_brown}{RGB}{140, 86, 75}
\definecolor{sb_pink}{RGB}{227, 119, 194}
\definecolor{sb_grey}{rgb}{0.4980392156862745, 0.4980392156862745, 0.4980392156862745}
\definecolor{sb_yellow}{rgb}{0.7372549019607844, 0.7411764705882353, 0.13333333333333333}
\definecolor{sb_colorblind_blue}{HTML}{0173b2}
\definecolor{sb_colorblind_orange}{HTML}{de8f05}
\definecolor{sb_colorblind_green}{HTML}{029e73}
\definecolor{sb_colorblind_red}{HTML}{d55e00}
\definecolor{sb_colorblind_purple}{HTML}{cc78bc}
\definecolor{sb_colorblind_brown}{HTML}{ca9161}
\definecolor{sb_colorblind_pink}{HTML}{fbafe4}
\definecolor{sb_colorblind_grey}{HTML}{949494}
\definecolor{sb_colorblind_yellow}{HTML}{ece133}
\definecolor{sb_colorblind_yellow}{HTML}{ece133}
\definecolor{paired_lightblue}{HTML}{a6cee3}
\definecolor{paired_darkblue}{HTML}{1f78b4}
\definecolor{paired_lightgreen}{HTML}{b2df8a}
\definecolor{paired_darkgreen}{HTML}{33a02c}
\definecolor{paired_pink}{HTML}{fb9a99}
\definecolor{paired_red}{HTML}{e31a1c}
\definecolor{paired_yellow}{HTML}{fdbf6f}
\definecolor{paired_orange}{HTML}{ff7f00}
\definecolor{paired_mauve}{HTML}{cab2d6}
\definecolor{paired_purple}{HTML}{6a3d9a}
\definecolor{paired_brown}{HTML}{b15928}
\definecolor{css4_dimgrey}{rgb}{0.4117647058823529, 0.4117647058823529, 0.4117647058823529}

\definecolor{lhl}{RGB}{234, 245, 250}
\definecolor{hl}{RGB}{205, 232, 248}

\newcommand{\fhl}[1]{\sethlcolor{sb_blue!25}\hl{#1}}
\newcommand{\shl}[1]{\sethlcolor{sb_blue!10}\hl{#1}}

\newcommand{\cblock}[1]{
 \begin{tikzpicture}
   [
   node/.style={square, minimum size=10mm, thick, line width=0pt},
   ]
   \node[fill={#1}] {};
 \end{tikzpicture}
}

\newcommand{\ccircle}[1]{
 \begin{tikzpicture}
   [
   node/.style={line width=0pt},
   ]
   \node[fill={#1}, circle, draw=white, scale=0.7] {};
 \end{tikzpicture}
}

\DeclareMathOperator*{\argmin}{argmin}

\makeatletter
\newcommand{\linebreakand}{%
  \end{@IEEEauthorhalign}
  \hfill\mbox{}\par
  \mbox{}\hfill\begin{@IEEEauthorhalign}
}
\makeatother

\def\BibTeX{{\rm B\kern-.05em{\sc i\kern-.025em b}\kern-.08em
    T\kern-.1667em\lower.7ex\hbox{E}\kern-.125emX}}
    
\begin{document}

\title{Generalizable Imitation Learning Through Pre-Trained Representations
}

\author{Wei-Di Chang, Francois Hogan, Scott Fujimoto, David Meger, and Gregory Dudek \thanks{Work done while Wei-Di Chang was interning at the Samsung AI Center Montr\'eal. Center for Intelligent Machines, McGill University. Corresponding author:  {\tt\footnotesize wchang@cim.mcgill.ca}.}
}

\maketitle

\begin{abstract}
In this paper, we leverage self-supervised vision transformer models and their emergent semantic abilities to improve the generalization abilities of imitation learning policies. We introduce DVK, an imitation learning algorithm that leverages rich pre-trained Visual Transformer patch-level embeddings to obtain better generalization when learning through demonstrations. Our learner sees the world by clustering appearance features into groups associated with semantic concepts, forming stable keypoints that generalize across a wide range of appearance variations and object types. We demonstrate how this representation enables generalized behaviour by evaluating imitation learning across a diverse dataset of object manipulation tasks. To facilitate further study of generalization in Imitation Learning, all of our code for the method and evaluation, as well as the dataset, is made available. 
\end{abstract}

\section{Introduction}

Once humans acquire a manipulation skill, they can immediately adapt it to unseen objects despite drastic variations in visual appearance and geometry. This is perhaps due to the human ability to identify operational concepts, such as seats or handles, which exist across many different object classes. 
Learned robotic behaviours, on the other hand, typically have difficulties generalizing beyond their training data. 

In this paper, we consider the setting of imitation-based manipulation policies for objects that are \textit{unseen} in the training set. Imitation Learning (IL) is a proven method for training complex robot behaviours from demonstrations. It offers better sample efficiency than learning behaviours from scratch through Reinforcement Learning (RL) and avoids the need to engineer a reward function~\cite{abbeel2004apprenticeship, ho2016generative, chang2022flow, florence2022implicit, chi2023diffusionpolicy, chang2023imitation}. Despite these successes, however, IL often relies on limited distribution shift, where the 
distribution of the
training dataset closely matches the final evaluation. 

Leveraging pre-trained visual representations is a common solution to sample efficiency and generalization. Prior work has shown their benefits in control and IL \cite{parisi2022unsurprising, nair2022r3m, xiao2022mvp}. Using pre-trained representations also comes with improved generalization that is achieved through training on larger more diverse training datasets. Finding the best Pre-trained Visual Representation (PVR), and pinpointing \emph{why} in particular it is the most effective as an artificial visual cortex for robotics has been of particular interest \cite{majumdar2023we, burns2023makes, dasari2023unbiased}. In contrast, little attention has been given to \emph{how} to use these PVRs best. 
The default approach to leverage them feeds a flat representation to the control component, an approach inherited from CNN-based architectures for classification \cite{simonyan2014very, he2016deep, krizhevsky2017imagenet}.
In the context of robotic manipulation, these flattened embeddings aim to encode the classification of objects present in the scene (what to grasp), and the location of related points of interest (where to grasp) \cite{shridhar2022cliport}. Yet most general-purpose PVRs have little guarantee about what they encode and lack interpretability. Some approaches aim to remedy this through the use of specific losses and training on tailored datasets \cite{nair2022r3m, xiao2022mvp, majumdar2023we, dasari2023unbiased}. We find that by explicitly tracking specific patch-level features of DINO Vision Transformers\cite{caron2021emerging}, we can significantly improve the generalization of policies learnt through imitation learning while remaining interpretable.

\input{tex_files/intro_figure_1line}

Our work advances generalization in IL through two main contributions.
We first introduce a new approach to better leverage ViT-based PVRs in imitation learning. Our approach,
DVK~(DINO ViT Keypoints), is an imitation learning algorithm based on behaviour cloning that leverages rich pre-trained DINO ViT patch-level embeddings to obtain better generalization when learning from demonstrations. Across the increasingly difficult cases of our benchmark, DVK shows improved transfer of the learnt manipulation policy to unseen objects, outperforming state-of-the-art IL approaches and PVRs designed for control. 

To better evaluate DVK, our second contribution is a benchmark to study object generalization. We build a set of benchmark tasks using the Google Scanned Objects dataset \cite{downs2022google} that tests the ability of IL algorithms for robotic manipulation to generalize under two challenging settings: within a class---when the manipulated objects change in both visual appearance and morphology, and across classes---with objects of vastly different appearance and morphology. 
Our benchmark is designed to study these generalization facets of IL independently and measure how transferable policies learnt through imitation are to novel, unseen objects. To facilitate reproducibility, all of our code, for both the method and the benchmark will be open-sourced.

\section{Related Work}

\textbf{IL for Manipulation.} 
Learning manipulation behaviours from demonstrations has been widely studied, where most methods broadly categorized as Inverse Reinforcement Learning \cite{abbeel2004apprenticeship} or Imitation Learning (IL).
Behaviour Cloning (BC) \cite{pomerleau1991efficient} remains widely used for its simplicity and effectiveness in real-world settings \cite{rahmatizadeh2018vision, florence2019self, robomimic2021}, despite known shortcomings \cite{ross2011reduction}. To improve spatial generalization and better handle system dynamics, various policy functional forms have been proposed \cite{florence2022implicit, chi2023diffusionpolicy}. %
A common technique is Spatial Softmax layers~\cite{levine16spatialsm} which are used to abstract CNN features into 2D image coordinates~\cite{wang2021generalization, kim2021gaze, yu2018one}. 
Our keypoint extraction approach is reminiscent of this image coordinate abstraction, but is adapted for ViTs and can track consistent semantic concepts across frames and rollouts. 

\textbf{Pre-trained Representations.} 
Prior work has shown that pre-trained embeddings can improve the performance and data efficiency of visual control policies~\cite{laskin2020curl, yarats2021image, yarats2021mastering, fujimoto2023sale, karamcheti2023language}.
Parisi et al.\ \cite{parisi2022unsurprising} find that representations pre-trained on general-purpose large-scale image datasets can outperform ground-truth state representations in transferability.

There are many self-supervised PVR approaches for visuomotor policies that leverage large-scale datasets. 
R3M~\cite{nair2022r3m} uses time-contrastive learning, video-language alignment, and a sparsity penalty with the Ego4D dataset \cite{grauman2022ego4d} for IL. %
MVP~\cite{xiao2022mvp} pre-trains a ViT-based Masked Auto-Encoder (MAE) on Imagenet~\cite{imagenet} to improve RL performance and transferability to novel objects. 
VC-1~\cite{majumdar2023we} compares many PVRs and adopts the same architecture as MVP, pre-training on ImageNet alongside  %
Ego4D. 
The Soup models \cite{dasari2023unbiased} are obtained by optimizing the embedding training data distribution by combining multiple large-scale datasets.
MVP, VC-1 and Soup all adopt the same MAE self-supervised loss and ViT architecture, and use the compact \texttt{[CLS]} image-level token as the output embedding. 
In contrast, we propose to use the dense patch-level embeddings from DINO pre-trained ViTs. These embeddings are better localized and, following our keypoint extraction method, better abstracted to avoid potential overfitting to specific feature values.

Consistent with prior work~\cite{burns2023makes}, we find DINO ViTs offer the best out-of-distribution performance. Aside from superior generalization, we leverage them for their part-based semantic abilities. \autoref{fig:objects} demonstrates this zero-shot ability, using just cosine similarity over the patch embeddings.

\textbf{Object Generalization in Control.} 
Domain randomization is a common method to address noise and shifts in visual sensing~\cite{tobin2017domain, sadeghi2018sim2real}. 
In this paper, we are motivated by the insight that humans abstract over not only appearance but also object shape and class. 
Object-abstraction in control has often relied on bounding boxes \cite{wang2019deep, devin2018deep, sieb2020graph, zhu2023viola} or pose estimation \cite{tremblay2018deep, du2021vision, wang2022geometry}. 
Dense Object Net \cite{florence2018dense} learns representations through self-supervised robot interaction and a dense correspondence loss, which enable keypoint localization on objects of the same class, but is constrained by its dependence on specialized collection of RGBD data for each of the manipulated objects using multiple camera views and %
limited to previously seen object classes. 
\cite{kulkarni2019unsupervised} also abstract images into keypoints for RL through their Transporter architecture and benefit from the increased efficiency of abstraction through keypoints, but rely on in-domain images for training, limiting its applicability to its training environment.

Unlike previous approaches, DVK generalizes to unseen objects and object classes by extracting visual object-part semantic concepts from expert demonstrations and tracking them as keypoints. 
It thus avoids the rigidity of bounding boxes and does not need supervision or environment interaction, leveraging instead the generality of embeddings trained on large-scale general datasets.

Previous benchmarks focus on policy generalization between different tasks \cite{yu2020meta}, manipulation of abstract 3D printed objects \cite{luo2024fmb}, intra-class generalization \cite{mu2021maniskill}, or large-scale variations in object appearance only, without variations in object morphology \cite{pumacay2024colosseum}. In contrast our benchmark also examines inter-class generalization of imitation learning policies over common household objects covering a wide range of objects spanning many classes,
 thus emphasizing differences in object morphology and appearance.

 \begin{figure}[t]
\centering
  \includegraphics[width=.35\textwidth]{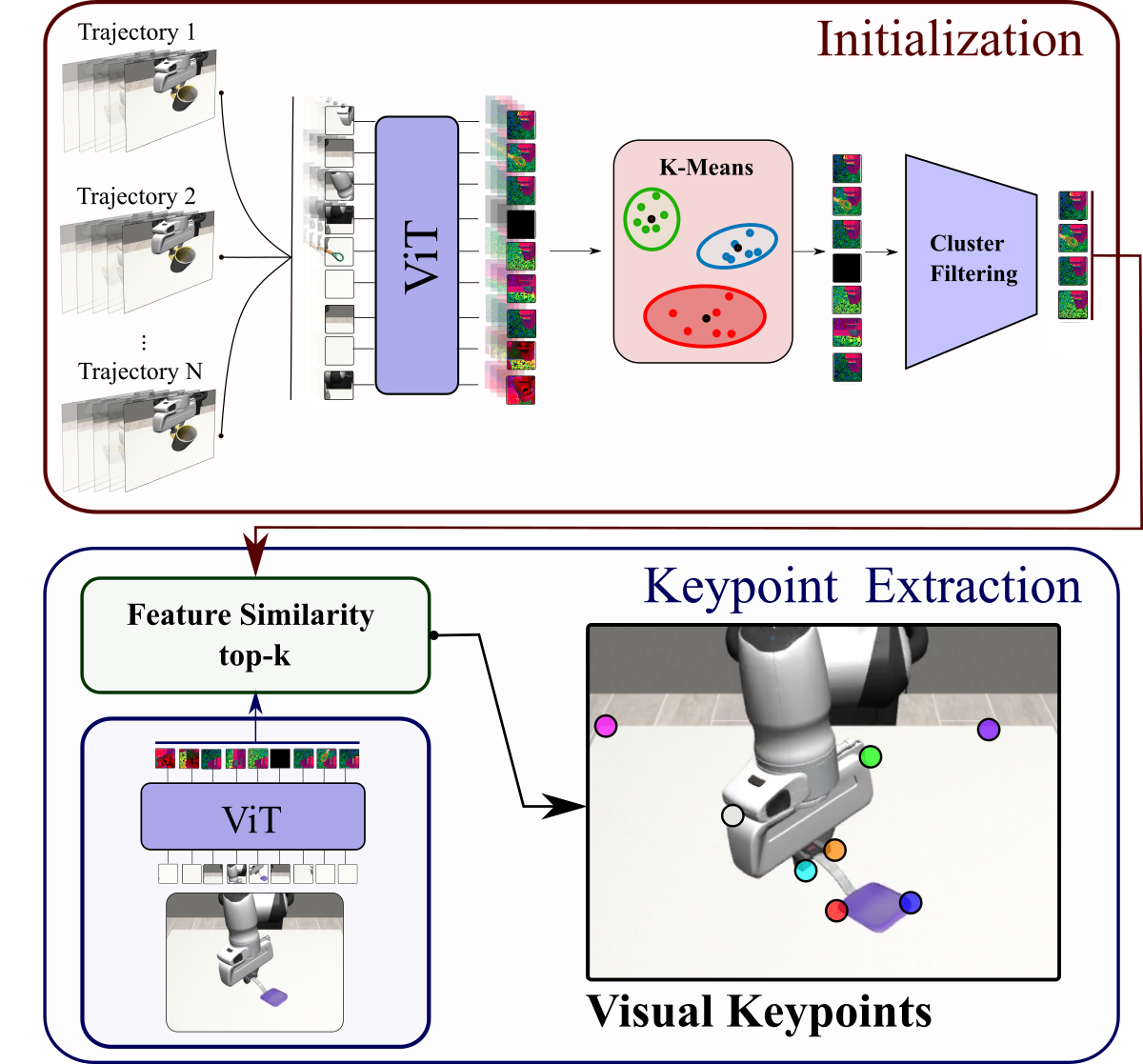}
\caption{\textbf{Overview of our feature extraction pipeline.} Our approach extracts reference DINO features through K-Means clustering that represent semantic concepts from the demonstration dataset. These reference features are used in downstream policy learning, to abstract images seen in rollouts into semantic visual keypoints.}
\label{fig:pipeline}
\vspace{-12pt}
\end{figure}

\section{Background} \label{sec:ViT}

\textbf{Vision Transformers.} 
The ViT architecture encodes a sequence of image patches into a sequence of patch embeddings using self-attention. From an image input as a grid of patches, each patch is flattened, linearly projected and processed by Multi-Head Self-Attention (MHSA).
Stacked MHSA and feedforward layers learn useful patch representations, maintaining feature map size across layers, unlike CNNs that reduce resolution via pooling layers. This results in finer feature maps, a beneficial aspect for robot policies.

ViTs features can be extracted by appending a \texttt{[CLS]} token to the input patch sequence and using the output \texttt{[CLS]} token as a global image embedding~\cite{kenton2019bert}. 
This compact image-level \texttt{[CLS]} embedding is often used for downstream tasks \cite{xiao2022mvp}, sidestepping the issue of the large output size of the full ViT embedding but at the cost of ignoring the dense patch-localized embeddings entirely. Another approach pools (average, global, or GeM pooling) and concatenates patch embeddings to the \texttt{[CLS]} token \cite{caron2021emerging}.
While some approaches to reduce ViTs' high-dimensional outputs exist~\cite{beyer2022better, caron2021emerging, renggli2022learning}, they are not ideal for generalization, as they use trainable layers that lose the generality of pre-trained features and sacrifice feature localization from averaging. 

\textbf{DINO Representations.}  
Self-\textbf{Di}stillation with \textbf{No} Labels (DINO)  \cite{caron2021emerging} representations are trained through self-supervision. 
In \autoref{fig:objects}, we show the zero-shot part-semantic abilities of DINO patch embeddings. We display each object's cosine similarity heatmap between its ViT DINO patch descriptors and that of a localized patch on reference object "Mug~A", marked with a red dot. 
ViT DINO extracts representations that locate similar semantic parts of objects on these images unseen in training. 

The `key' token patch embeddings~$h$ are extracted from the last layer of DINO ViT, forming the $N \times N$ grid:
    $H = \{h_{(x_i,y_j)}\}_{i=1,j=1}^{N,N}$.
Our method leverages these patch-level embeddings, with minimal processing or further learning on top of the frozen representation, preserving both generality and patch-specific localization to achieve IL policies that generalize across object classes. 

\textbf{Behaviour Cloning.} 
Behaviour Cloning (BC) learns a policy $\pi_\theta$ mapping state $s$ to action $a$: $\pi_\theta(s) = a$, by minimizing a regression loss on dataset $D_E$ of expert demonstration state-action pairs $(s_E, a_E)$:
\begin{equation}
\argmin_\theta \mathbb{E}_{(s_E, a_E) \sim D_E} \norm {\pi(s_E) - a_E}^2 .    
\label{eq:bc}
\end{equation}
For image inputs, the visual backbone encodes the image, and the flattened image representation is concatenated with the rest of the state variables for input to the policy.

 \begin{figure*}[t]
\centering
  \includegraphics[width=1.\textwidth]{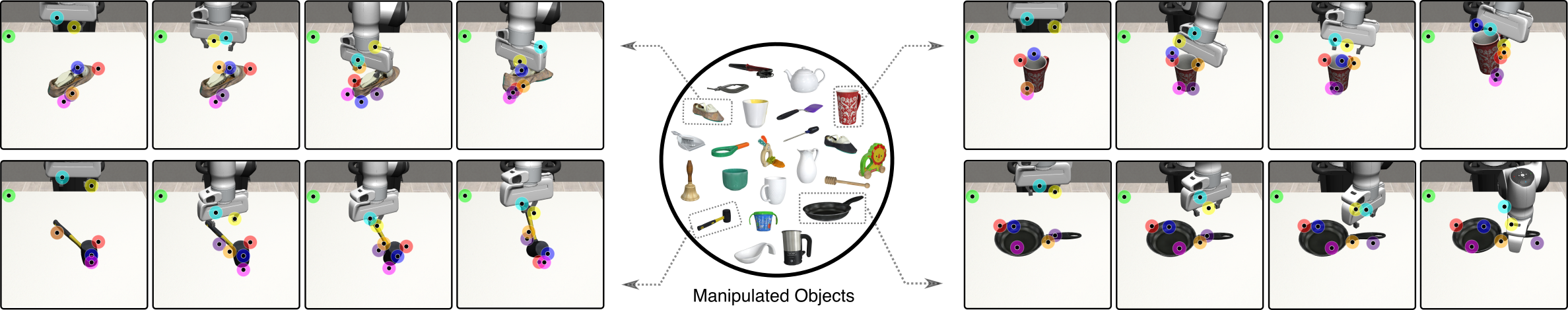}
\caption{\textbf{Example grasping rollouts.} Our keypoints (represented by colored circles) track semantic concepts extracted from the expert demonstrations. In the shown rollouts, we display keypoints that track parts of objects such as their handles, tips and edges, parts of the end-effector, and static elements of the workspace such as the corner of the table.}
\label{fig:graph_rollout}
\vspace{-15pt}
\end{figure*}

\section{Method}
To benefit from the localized semantic descriptors of DINO, our framework, DVK (DINO ViT Keypoints), processes image features into 2D keypoints before feeding them to the BC policy. The first step is to initialize reference features through clustering on expert demonstration images. These initialized reference features are then used to extract keypoints from individual images in both policy training and testing. A schematic of our pipeline is shown in \autoref{fig:pipeline}.

Abstracting the dense visual representation to localized keypoints offers two key advantages compared to DINO patch embeddings, i) it significantly reduces the input size for the imitation policy, ii) it eliminates overfitting to specific DINO features for the downstream policy while preserving the generality of the pre-trained backbone features.

\textbf{Reference Features Initialization.}
To identify pertinent keypoints, DINO patch embeddings from expert demonstrations are clustered into sets representing semantic concepts in the demonstration trajectories. This is performed as follows: 
\begin{enumerate}
    \item We first extract the $N\times N$ grid of DINO image patch descriptors from all $J$ images in the expert demonstrations, as described in section \ref{sec:ViT}.
    \item The resulting $J\times N \times N$ patch embeddings are treated as a bag of features, disregarding their image location, sequence location, or associated object.
    \item K-Means clustering is then applied to the embeddings, resulting in $M$ candidate clusters with $M$ corresponding cluster centroids. 
\end{enumerate}

To keep only salient clusters found in most demonstration images, we use the voting system and saliency defined by Amir et al.~\cite{amir2021deep}. %
Formally, let $a_i^I$ denote the mean \texttt{[CLS]} token attention of the MHSA heads in the last layer of the ViT for patch $i$ in image $I$. Let $S_j^I$ be the set of all patches in image $I$ belonging to the $j$-th cluster candidate. The saliency for $S_j^I$ is the average attention value of patches~in~$S_j^I$: $\mathrm{sal} (S_j^I) = \frac{1}{|S_j^I|} \sum_{i \in S_j^I} a_i^I$.
Iterating over all images, a vote is cast by image $I$ for cluster candidate $C_j,\, 0<j<M$ if the saliency of this cluster in the image exceeds a threshold value $\tau$: $\mathrm{sal} (S_j^I) > \tau$. Ranking the cluster candidates by votes, we keep a portion $m \leq M$ of the most voted for clusters, saving their reference centroids as the set $F = \{f_0, \cdots, f_{m-1}\}$.

\textbf{Keypoint Extraction.}
After the initialization in the previous step, keypoint coordinates are computed as the locations of the closest feature patch relative to each cluster centroid.

For each frame input~$I$ to the policy, we use DINO ViT~\cite{caron2021emerging} and extract the `key' patch embeddings~$h$ from its last layer, forming the grid~$H = \{h_{(x_i,y_j)}\}_{i=1,j=1}^{N,N}$. From $H$ we compute an output vector composed of $m$ image keypoint coordinates $\mathbf{K}(I) = {(x_k, y_k)}_{k=1}^m$, corresponding to the coordinates of the closest patch DINO feature in the current frame to each of the reference centroids in the set $F$: 
\begin{equation}
    (x_k, y_k) = \argmin_{(x_i, y_j)} \frac{f_k \cdot h_{(x_i,y_j)}}{||f_k||~||h_{(x_i,y_j)}||}.    
\end{equation}
This output vector of image keypoint coordinates $\mathbf{K}(I)$ forms the visual representation input to the policy head. %
This procedure allows DVK to track a set of representative semantic concepts, extracted from the demonstration image dataset by DINO. If a concept is not present in the current frame, DVK outputs the location of the closest concept. 

In \autoref{fig:graph_rollout} we provide an example of our extracted keypoints, which track various elements in the rollouts, such as specific parts of the gripper, objects (handle, tip), or fixtures of the environment (e.g.\ the corner of the table).

\textbf{Policy Learning.}
Once the keypoints vector $\mathbf{K}(I)$ is extracted, we minimize the BC loss:  %
\begin{equation}
\argmin_\theta \mathbb{E}_{(I, s, a) \sim D_E} \norm {\pi(s, \mathbf{K}(I)) - a}^2 .    
\end{equation}
with the input state composed of the visual input abstracted as keypoints, $\mathbf{K}(I)$, and proprioceptive state elements, end-effector position, and rotations captured by the expert states~$s$. During evaluation, keypoints are extracted similarly using the current image in the rollout. 

\begin{figure}[ht]
\centering
\vspace{8pt}
\begin{overpic}[width=0.5\textwidth,keepaspectratio]{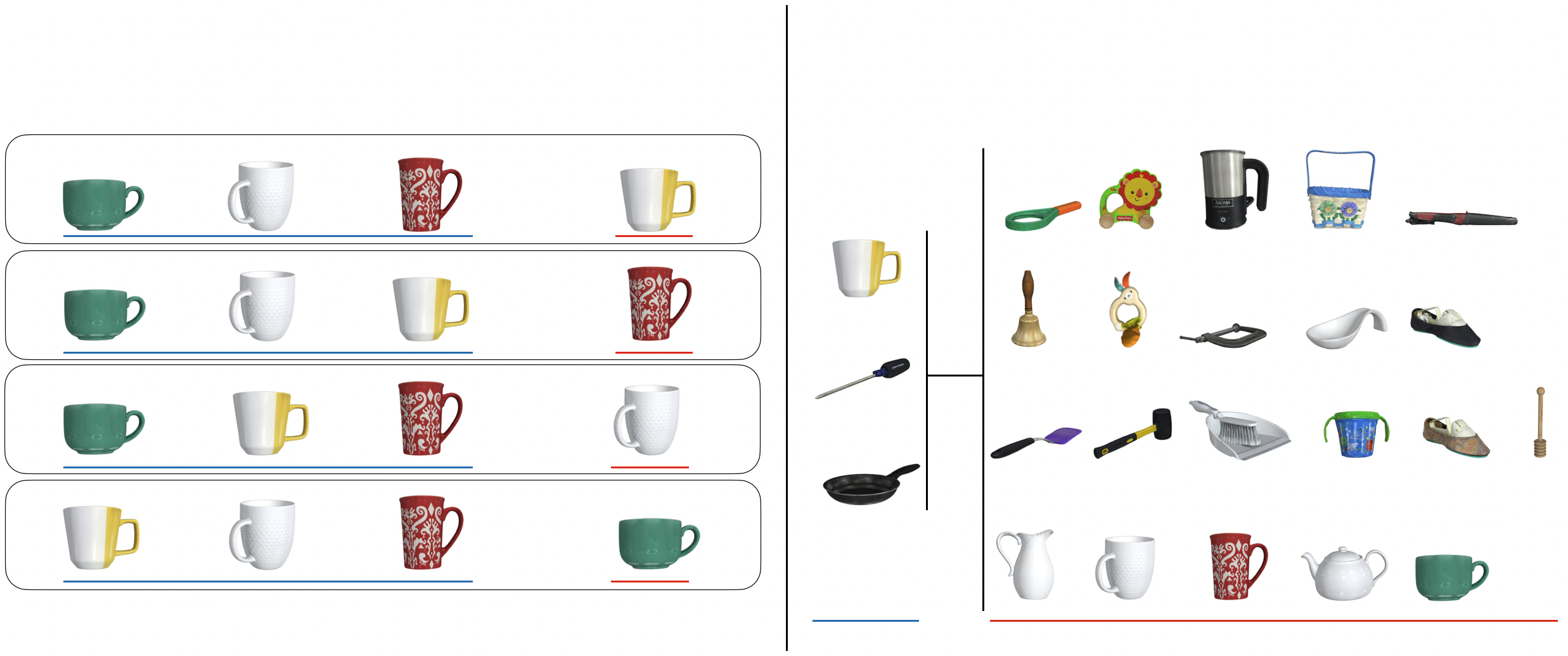}
\put(2,35){\shortstack{\footnotesize \textbf{Intra-Class Generalization} \\\footnotesize Single class\\\footnotesize Various shapes and appearances}}
\put(55,35){\shortstack{\footnotesize \textbf{Inter-Class Generalization} \\\footnotesize Multi-class\\\footnotesize Diverse shapes and appearances}}
\put(13,-1){\footnotesize Train}
\put(39.5,-1){\footnotesize Test}
\put(52,-1){\footnotesize Train}
\put(80,-1){\footnotesize Test}
\put(1.5, 28.5){\footnotesize 1}
\put(1.5, 21){\footnotesize 2}
\put(1.5, 13.75){\footnotesize 3}
\put(1.5, 6.5){\footnotesize 4}
\end{overpic}
\vspace{-8pt}
\caption{\textbf{Intra-Class and Inter-Class Generalization Benchmark.} Our grasping benchmark examines the intra-class generalization abilities of policies through four-fold cross-validation on the Mug class, and inter-class generalization through transfer from 3 objects to 21 unseen objects of various classes.} \label{fig:dataset} 
\vspace{-8pt}
\end{figure}

\section{Grasping Generalization Benchmark}
To evaluate the generalization abilities of manipulation policies, we develop a benchmark using 24 objects from the Google Scanned Objects Dataset \cite{downs2022google}, a dataset of physically faithful and photorealistic 3D-scanned household objects made for manipulation. Example objects are displayed in \autoref{fig:objects}, and the full set, in \autoref{fig:dataset}. The subset was chosen to represent a wide range of manipulation difficulty and a range of potential for policy transfer between the objects. 

Our benchmark studies how changes to the perceptual backbone of the visuomotor policy can improve its object generalization. We focus on two key facets: \textit{intra-class} generalization, where the agent is trained and tested on objects of the same class, with variations in shape and appearance, and \textit{inter-class} generalization, where the agent is tested on entirely new object classes unseen during training.

\textbf{Task.} Each task involves grasping and lifting an object from a table using a Panda manipulator, simulated in Robosuite~\cite{robosuite2020}. The agent has access to the proprioceptive state of the arm but relies on camera inputs for object perception. Consistent with prior work \cite{robomimic2021}, all methods operate using an end-effector positional operational space controller~\cite{khatib1987unified,peters2006learning}. We focus on a single basic task to best isolate the impact of the visual backbone from the role of the controller.

\begin{figure}[t]
\centering
  \includegraphics[width=.3\textwidth]{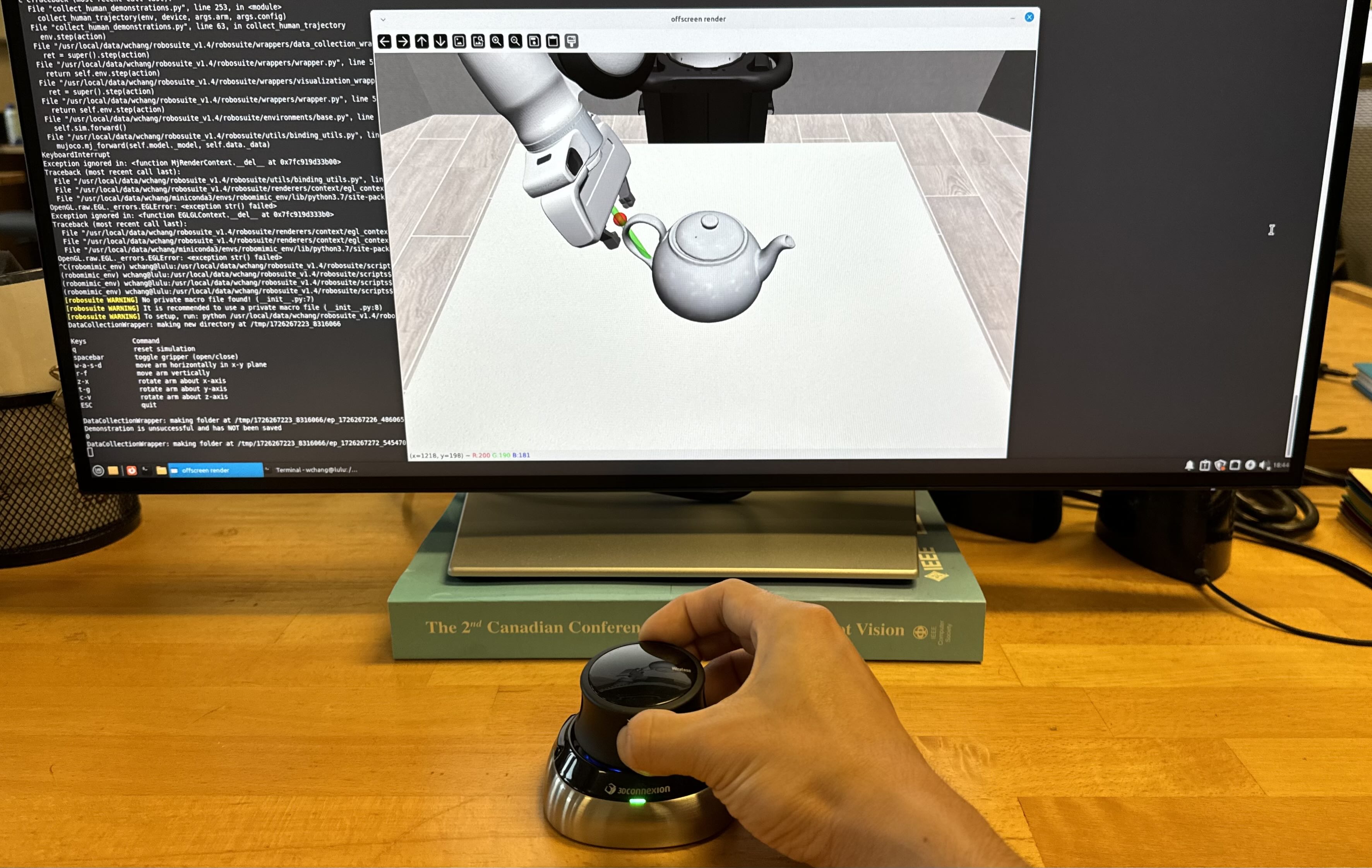}\\
 \caption{\textbf{Demonstration collection setup using a spacemouse in Robosuite.} Our benchmark consists of 60 demonstration trajectories collected through human teleoperation for each training object.}
 \vspace{-18pt}
\label{fig:spacemouse}
\end{figure}

\textbf{Demonstrations.} Expert demonstrations are collected by an operator through teleoperation using a spacemouse in Robosuite. The training dataset 
for each benchmark 
consists of $60$ demonstration trajectories for each of the objects in the dataset, and will be released alongside our code. We show in \autoref{fig:spacemouse} our demonstration collection setup.

Once trained, the policies are tested with unseen objects, performing the same task shown in training.

\textbf{Intra-Class Generalization Benchmark.} We use all four mugs available in the dataset~\cite{downs2022google} as training objects, each with various appearances and shapes but belonging to the same object class. The demonstrated task consists of grasping the mug by its \emph{handle} and lifting it.
We test generalization through a leave-one-out cross-validation scheme, forming four splits of three training objects and one test object, %
of the same class but unseen shape and appearance ~(\autoref{fig:dataset}).

\textbf{Inter-Class Generalization Benchmark.} 
To assess generalization performance across classes, 
we train on three diverse objects: a yellow mug, a pan, and a screwdriver. The remaining 21 objects are used as a test set with unseen shape and appearance~(\autoref{fig:dataset}), presenting a wide range of difficult manipulation tasks 
with objects: of the same classes with different morphologies (mugs), of the same semantic parent class (tools, liquid containers) that share similar design features (straight handles and side handles), and objects of entirely different classes with differing degrees of handle-like appendages (basket, toys, shoes with shoetree).

\input{tex_files/intra_table_highlight}

\section{Results}

Our experiments examine the extent to which robotic policies trained through BC can generalize to changes in object shapes and appearances. 
We first focus on the ability of policies to generalize in the intra-class case, within the class of objects seen in training but with unseen shapes and appearances, followed by the inter-class generalization benchmark, with objects of various classes, all of shapes and appearances unseen in the demonstrations.

\textbf{Baselines.}
We use state-of-the-art baselines in manipulation \cite{robomimic2021} and pre-training representation methods for IL \cite{nair2022r3m,xiao2022mvp}. 
Like DVK, images are processed by encoders and passed to the BC policy alongside proprioceptive state elements. 

Scratch methods use randomly initialized visual encoders trained from scratch, while SL methods use frozen visual encoders pre-trained on Imagenet \cite{imagenet} with a supervised classification loss \cite{he2016deep, robomimic2021}. We also introduce DINO baselines that use a frozen visual encoder pre-trained on Imagenet with a DINO self-supervised loss, forming the ResNet \cite{he2016deep} (CNN) counterpart to our approach.
For each of these feature encodings, we use a standard Multi-Layer Perceptron (MLP) policy head, as well as a variant using a Recurrent Neural Network (RNN) into a Gaussian Mixture Model policy head as introduced in \cite{robomimic2021}. Each of these baselines uses a Spatial Softmax layer at the encoder output for its performance and generalization benefits~\cite{levine16spatialsm} and a ResNet-50 \cite{he2016deep} backbone for fair comparison\footnotemark[1]. Scratch and SL methods 
achieve state-of-the-art performance in manipulation~\cite{robomimic2021}, and DINO-based methods provide a strong baseline to DVK.  

\footnotetext[1]{Closest ResNet variant in number of weights relative to the variant of ViT used in DVK, ViT-S (21M weights for ViT-S vs 23M for ResNet-50).}

We also evaluate state-of-the-art pre-trained representations from R3M~\cite{nair2022r3m} (Resnet-50),
MVP~\cite{xiao2022mvp} (ViT), VC-1~\cite{majumdar2023we} (ViT), and Soup~\cite{dasari2023unbiased} (ViT), which all have been shown to improve 
IL and control performance.

\begin{figure}[t]
\centering
  \includegraphics[width=.4\textwidth]{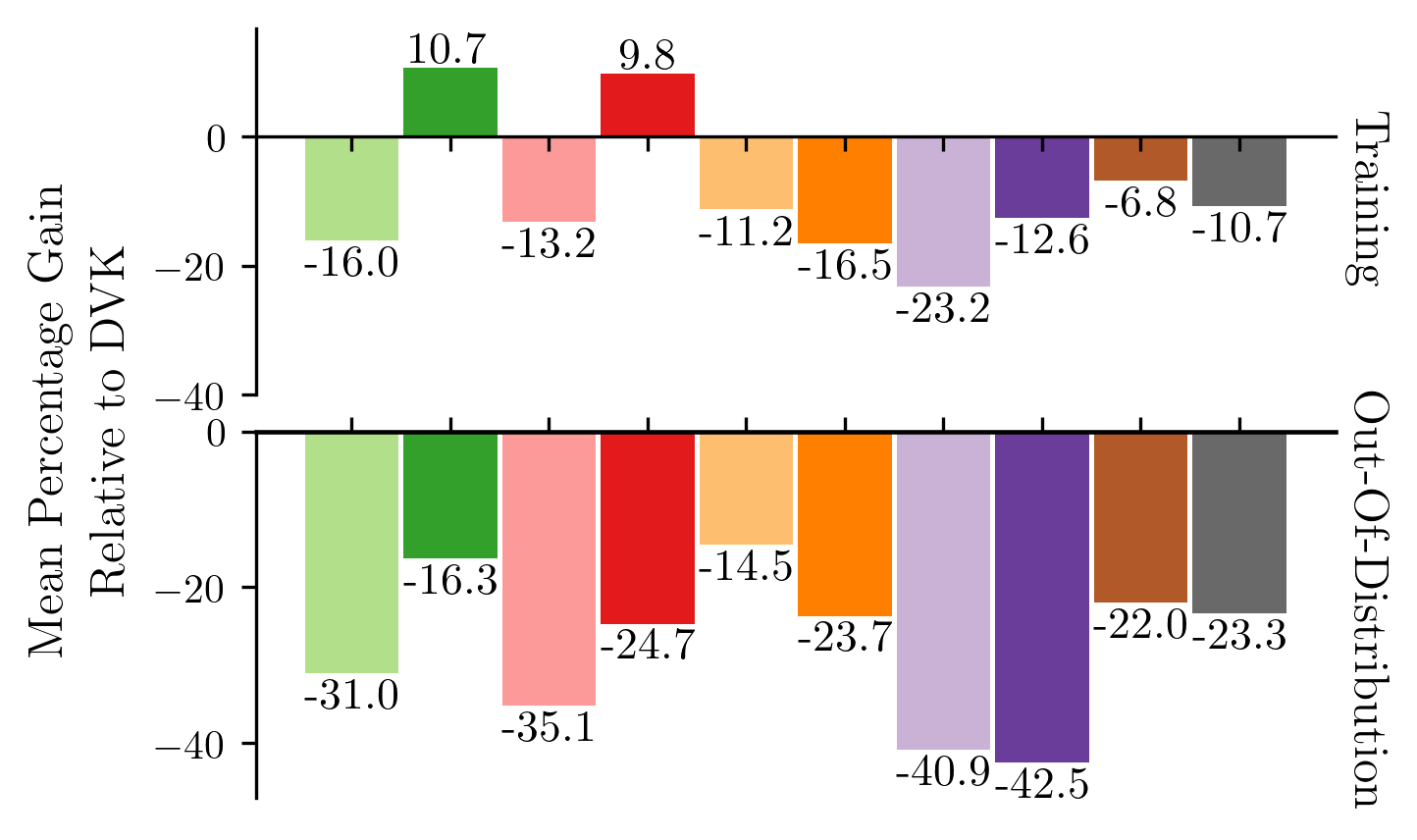}\\
\footnotesize \cblock{paired_lightgreen} DINO-MLP \cblock{paired_darkgreen} DINO-RNN  \cblock{paired_pink} SL-MLP \cblock{paired_red} SL-RNN \\ \cblock{paired_yellow} Scratch-MLP \cblock{paired_orange} Scratch-RNN \cblock{paired_mauve} MVP \cblock{paired_purple} R3M \cblock{paired_brown} VC-1 \cblock{css4_dimgrey} Soup
 \caption{\textbf{Training and out-of-distribution performance \emph{relative to DVK} on the intra-class generalization experiment.} Our method, DVK consistently improves the average success rate by a minimum of $15\%$ on previously unseen objects.}
\label{fig:exp1_delta}
\vspace{-12pt}
\end{figure}

\textbf{Evaluation.} In alignment with the methodology of~\cite{robomimic2021} we train the visuomotor policies for 600 epochs, rolling out in the training environment every 20 epochs and use the weights of the policy that performs best on the training environment for evaluation. Our evaluation then consists of 50 rollouts for each test environment, rotating the test object in the environment to cover 360 degrees.

\subsection{Intra-Class Generalization}

We first study the capacity of BC policies to generalize to objects within the same class as the expert demonstrations while varying shapes and appearances, testing the ability of trained policies to abstract visual semantics from the training dataset to novel, unseen objects of the same class. 

While easier than inter-class generalization, the intra-class setting is challenging as the demonstrations only show a very specific behaviour that lifts the mugs from their handle, making it a behaviour that requires a degree of precision, with variations based on the width, height, and handle shape of the mug in question. Results are reported in \autoref{tab:exp1} and \autoref{fig:exp1_delta} for each train-validation split.

\textbf{Discussion.} 
While DINO-RNN performs the best on training dataset, DVK outperforms all baselines on the test set, showing superior out-of-distribution generalization of our extracted keypoints. 
Vanilla BC~(Scratch-MLP) performs surprisingly well in the out-of-distribution context, on par with DINO-RNN, showing that learning a representation from scratch can be competitive with other pre-trained CNN-based approaches, but falls 15$\%$ behind DVK in average success rate. 
The PVR-based baselines, MVP, R3M, VC-1, and Soup perform worse than most other baselines on unseen objects, with VC-1 performing the best out of the four.

\input{tex_files/inter_table_new}

\subsection{Inter-Class Generalization}

\begin{figure}[h]
\centering
  \includegraphics[width=.4\textwidth]{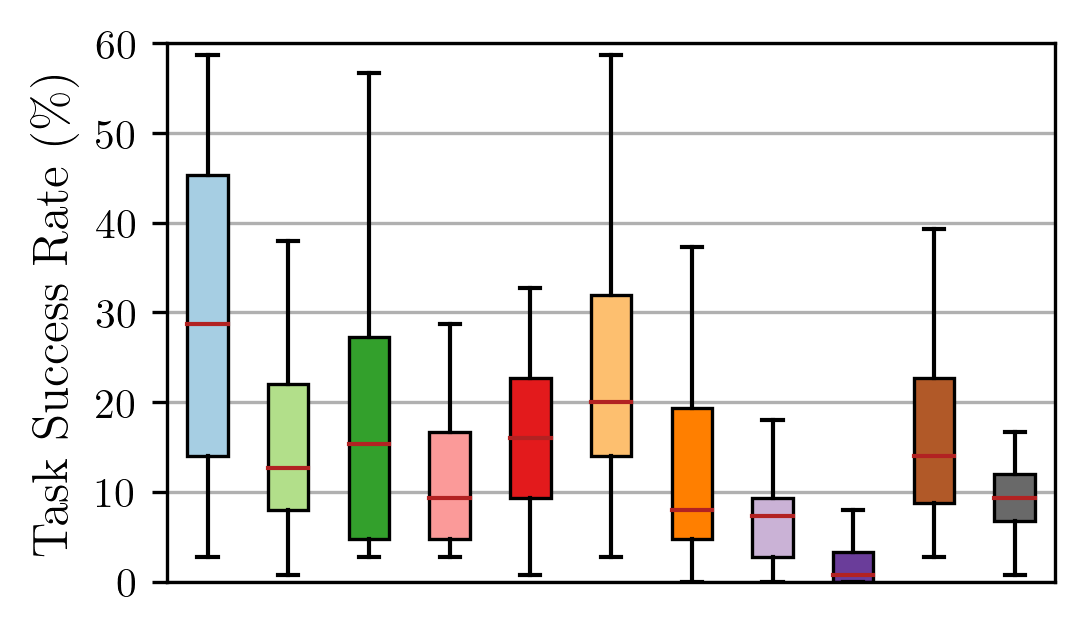}\\
\footnotesize \cblock{paired_lightblue} DVK \cblock{paired_lightgreen} DINO-MLP \cblock{paired_darkgreen} DINO-RNN  \cblock{paired_pink} SL-MLP \cblock{paired_red} SL-RNN  \\ \cblock{paired_yellow} Scratch-MLP \cblock{paired_orange} Scratch-RNN \cblock{paired_mauve} MVP \cblock{paired_purple} R3M \cblock{paired_brown} VC-1 \cblock{css4_dimgrey} Soup
\caption{\textbf{Box plot of task success rates for DVK and 10 baselines on the 21 out-of-distribution objects of the inter-class generalization experiment.} 
Whiskers represent 5th and 95th percentile, box bottom and top represent 25th and 75th percentile, and the red bar, the median.
DVK leads the studied methods, performing best on the unseen test set objects.}
\label{fig:ranking}
\vspace{-15pt}
\end{figure}

The inter-class benchmark poses a more difficult challenge than the previous setting as the test objects may no longer share physical characteristics or appearance with the objects found in the training dataset of demonstrations. Our results are reported in \autoref{tab:exp3} and aggregated in \autoref{fig:ranking}.

\textbf{Discussion.} We find that DVK outperforms all other methods on this difficult test set and DINO-RNN leads performance on the seen objects, showing the positive effect of using DINO features. 
DVK, as shown in \autoref{fig:ranking}, obtains a better median task success rate on test objects than competing methods.
It outperforms all methods on 38$\%$ of the unseen test objects, and places second best or third on most other objects (DVK is top-3 on 86$\%$ of test objects). 
BC from scratch (Scratch-MLP) performs second best in 
out-of-distribution tasks, showing that with a diverse set of training objects, BC is sufficient to learn representations that transfer well between objects. 

\subsection{Ablation Studies}
To validate the feature extraction method used in DVK, we evaluate alternative ways commonly used for feature extraction with ViTs: using only the \texttt{[CLS]} token (as done in MVP, VC-1, Soup), and using the \texttt{[CLS]} token concatenated with pooled patch embeddings. We also ablate the main hyperparameter of DVK, which is the number of candidate clusters used in the K-Means step. 

Removing the keypoint abstraction of DVK leads to large drops in train and test performance. Increasing the number of reference features in DVK leads to marginal gains in performance but also significantly increases the memory requirement of the K-Means clustering step.

\input{tex_files/ablation_table} 
\section{Conclusion}

In this paper, we study the generalization properties of imitation learning systems. We introduce a benchmark derived from the Google Scanned Objects dataset~\cite{downs2022google} and the Panda manipulator simulation in Robosuite~\cite{robosuite2020}, alongside a dataset of expert demonstrations. This benchmark evaluates 
intra-class variations such as color or shape, as well as major inter-class semantic differences between training and test data. 

To tackle our benchmark, we present DVK, an imitation learning framework that leverages dense DINO features abstracted into image keypoints as the visual representation for visual behaviour cloning. %
We find that DVK consistently handles out-of-distribution generalization better than existing state-of-the-art methods. %

DVK highlights the potential of stable semantic representations to ground robotic behaviour. Our keypoints track critical features on all of the gripper, target object, and background, enabling zero-shot adaptation to unseen objects--a step towards robots with more human-like manipulation. %

\bibliographystyle{IEEEtran}
\bibliography{main}
\end{document}

%% file: tex_files/intro_figure_1line.tex
\begin{figure}[t!]
\setlength\tabcolsep{3pt}

{$\hspace{8pt}$ \small Reference patch} 
\vspace{-2pt}

{\raggedright $\hspace{25pt} \downarrow$}
\vspace{-10pt}

\centering{
\begin{tabular}{ccccccccccccc}
 \begin{subfigure}{.06\textwidth}
 \includegraphics[width=\linewidth]{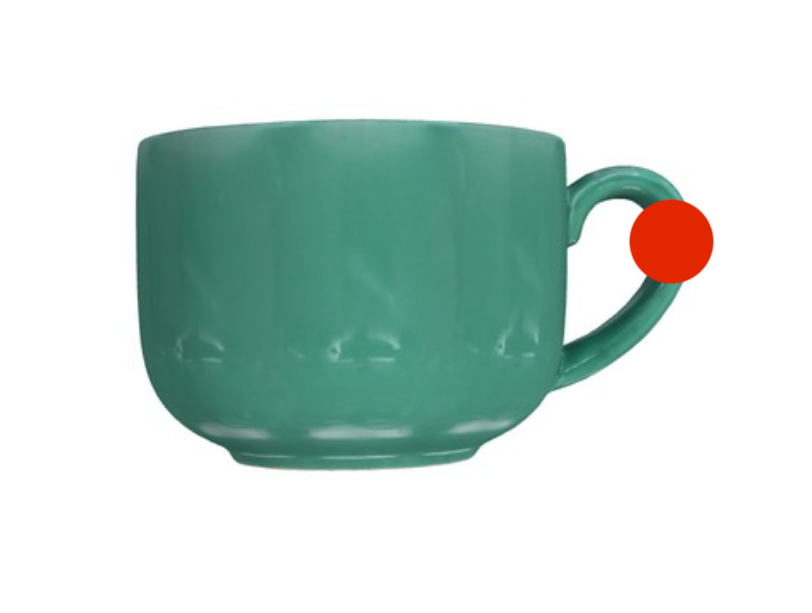}
  \end{subfigure} &
 \begin{subfigure}{.05\textwidth}
 \includegraphics[width=\linewidth]{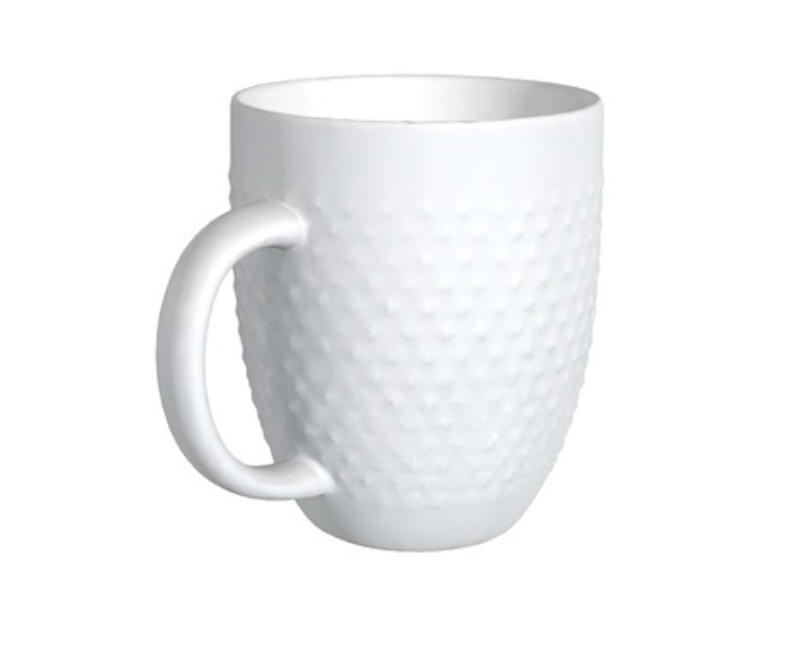}
  \end{subfigure}&
 \begin{subfigure}{.06\textwidth}
 \includegraphics[width=\linewidth]{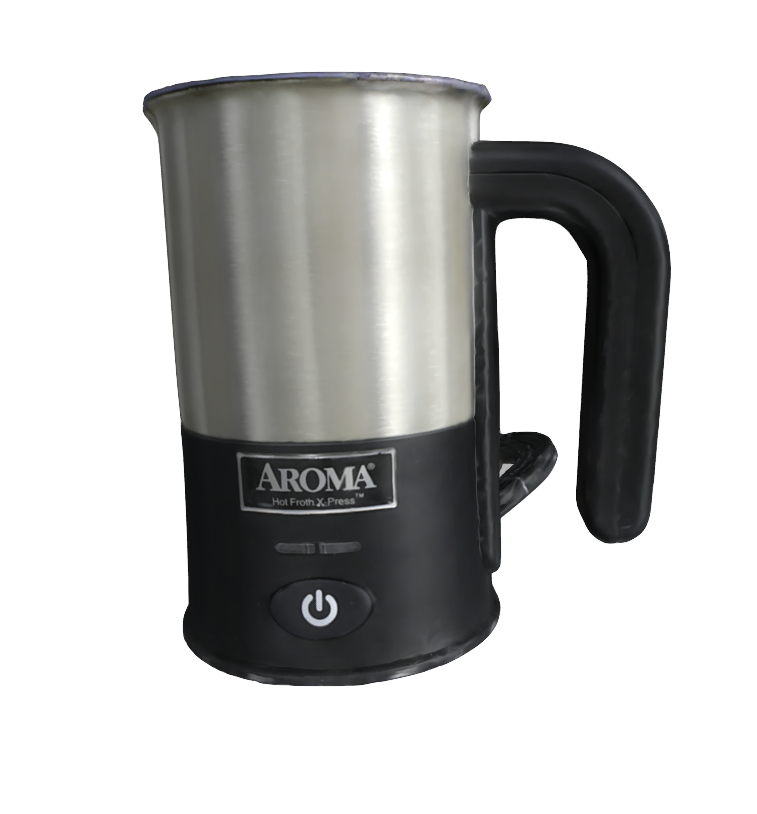}
  \end{subfigure}&
\begin{subfigure}{.06\textwidth}
 \includegraphics[width=\linewidth]{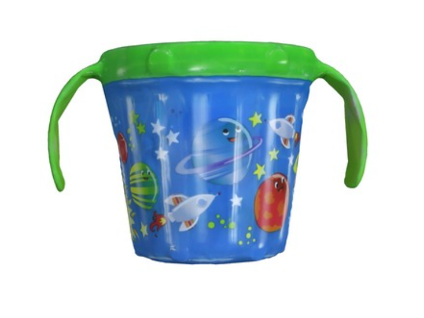}
  \end{subfigure}&
\begin{subfigure}{.06\textwidth}
 \includegraphics[width=\linewidth]{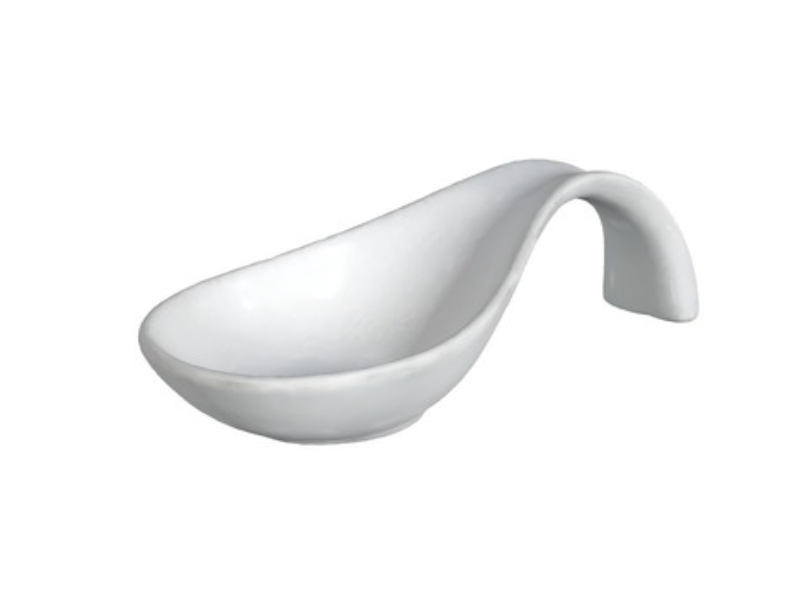}
\end{subfigure}&
\begin{subfigure}{.06\textwidth}
 \includegraphics[width=\linewidth]{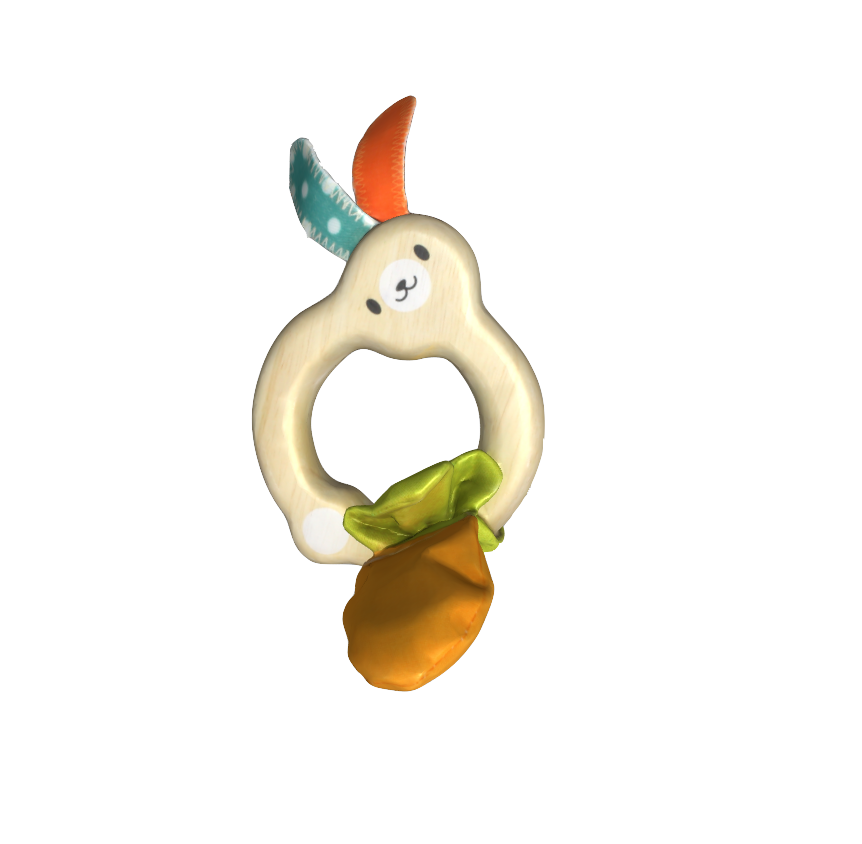}
\end{subfigure}&
   \multirow{3}{*}[2em]{
   \begin{subfigure}{.05\textwidth}
 \includegraphics[width=.8\linewidth]{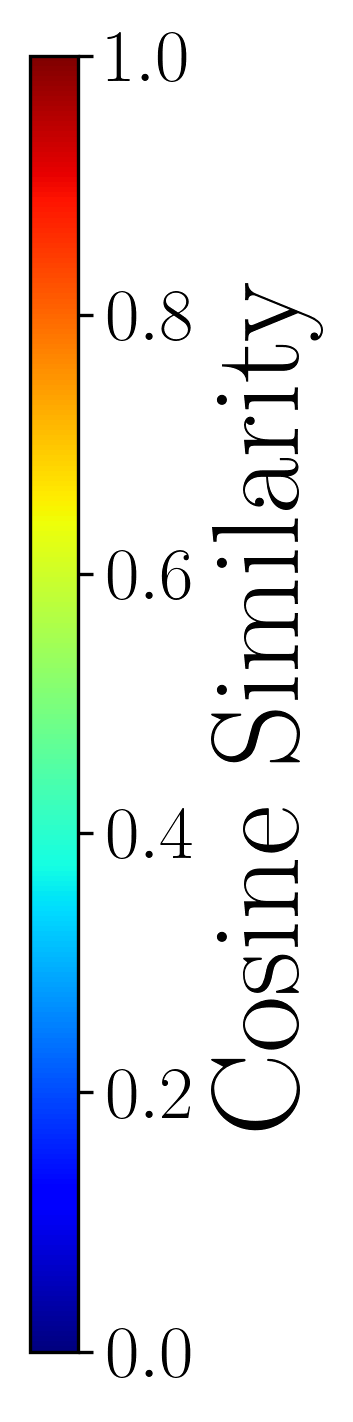}
  \end{subfigure}}
  \\

\begin{subfigure}{.06\textwidth}
$\quad\,\uparrow$ \\ \small{Reference Image}
  \end{subfigure}&
\begin{subfigure}{.06\textwidth}
 \includegraphics[width=\linewidth]{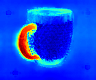}
  \end{subfigure}&
 \begin{subfigure}{.06\textwidth}
 \includegraphics[width=\linewidth]{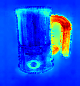}
  \end{subfigure}&
 \begin{subfigure}{.06\textwidth}
 \includegraphics[width=\linewidth]{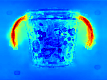}
\end{subfigure} &
\begin{subfigure}{.06\textwidth}
 \includegraphics[width=\linewidth]{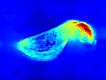}
\end{subfigure}&
\begin{subfigure}{.06\textwidth}
   \includegraphics[width=\linewidth]{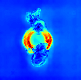}
\end{subfigure}&

    \\
 \small Mug A & 
  \small Mug B &
  \small Coffee &
  \small Kidcup &
  \small Spoonrest &
  \small Rattle &

\end{tabular}
}

\caption[]{\textbf{DINO ViT embeddings encode zero-shot fine-grained part-semantic information.} Cosine Similarity heatmap between DINO image patch embeddings of various household objects and reference patch {\hspace{-5pt} \small \protect\ccircle{red} \hspace{-5pt}}, all previously unseen. 6 of the 24 objects from the Google Scanned Objects dataset%
\cite{downs2022google} used in our manipulation experiments to test the generalization abilities of IL policies.}
\label{fig:objects}
\vspace{-16pt}
\end{figure}

%% file: tex_files/intra_table_highlight.tex
\begin{table*}[t]
\centering
\scriptsize
\setlength{\tabcolsep}{2pt}
\centering
\begin{tabularx}{\textwidth}{llXXXXXXXXXXX}
\toprule
\multicolumn{2}{l}{Feature Extractor Method} & \multicolumn{1}{l}{DVK} & \multicolumn{2}{l}{DINO} & \multicolumn{2}{l}{SL} & \multicolumn{2}{l}{Scratch} & \multicolumn{1}{l}{MVP} & \multicolumn{1}{l}{R3M} & \multicolumn{1}{l}{VC-1} & \multicolumn{1}{l}{Soup}\\
\cmidrule(r){3-3} \cmidrule(r){4-5} \cmidrule(r){6-7} \cmidrule(r){8-9}
\cmidrule(r){10-10} \cmidrule{11-11} \cmidrule{12-12} \cmidrule{13-13}
\multicolumn{2}{l}{Policy Network} & MLP & MLP & RNN & MLP & RNN & MLP & RNN & MLP & MLP & MLP & MLP\\
          \midrule
Train     & Mug (A, B, C)                  & 56.9 \textcolor{gray}{$\pm$ 6.3}                  & 43.1 \textcolor{gray}{$\pm$ 7.4}             & 68.0 \textcolor{gray}{$\pm$ 4.1}                  & 43.3 \textcolor{gray}{$\pm$ 1.4}          & \fhl{69.3}\textcolor{gray}{\fhl{ $\pm$ 4.7}                    } & 44.4 \textcolor{gray}{$\pm$ 2.6}                & 42.4 \textcolor{gray}{$\pm$ 5.2}      & 35.6 \textcolor{gray}{$\pm$ 4.4}              & 46.7 \textcolor{gray}{$\pm$ 2.9}                & 50.4 \textcolor{gray}{$\pm$ 2.9}  &  48.2 \textcolor{gray}{$\pm$ 2.3}\\
Test      & Mug D                                 & \fhl{63.3}\textcolor{gray}{\fhl{ $\pm$ 9.4}         } & 7.3 \textcolor{gray}{$\pm$ 1.9}              & 16.7 \textcolor{gray}{$\pm$ 6.6}                  & 16.7 \textcolor{gray}{$\pm$ 6.2}          & 31.3 \textcolor{gray}{$\pm$ 5.0}                             & 47.3 \textcolor{gray}{$\pm$ 3.4}         & 37.3 \textcolor{gray}{$\pm$ 5.7}      & 6.0 \textcolor{gray}{$\pm$ 2.8}               & 6.0 \textcolor{gray}{$\pm$ 4.9}                 & 37.3 \textcolor{gray}{$\pm$ 2.5}  &  26.0 \textcolor{gray}{$\pm$ 4.3}\\  \midrule
Train     & Mug (A, B, D)                 & 62.0 \textcolor{gray}{$\pm$ 2.4}                  & 41.1 \textcolor{gray}{$\pm$ 4.6}             & \fhl{72.4}\textcolor{gray}{\fhl{ $\pm$ 1.7}         } & 50.2 \textcolor{gray}{$\pm$ 5.1}          & 70.0 \textcolor{gray}{$\pm$ 6.4}                             & 51.3 \textcolor{gray}{$\pm$ 2.5}                 & 45.1 \textcolor{gray}{$\pm$ 6.5}      & 37.8 \textcolor{gray}{$\pm$ 7.0}              & 48.9 \textcolor{gray}{$\pm$ 4.6}                & 57.3 \textcolor{gray}{$\pm$ 4.3}  &  50.4 \textcolor{gray}{$\pm$ 1.3}\\
Test      & Mug C                                  & \fhl{46.7}\textcolor{gray}{\fhl{ $\pm$ 9.0}         } & 26.0 \textcolor{gray}{$\pm$ 5.9}             & 40.0 \textcolor{gray}{$\pm$ 4.9}                  & 12.7 \textcolor{gray}{$\pm$ 8.2}          & 24.0 \textcolor{gray}{$\pm$ 0.0}                             & 40.0 \textcolor{gray}{$\pm$ 4.3}        & 29.3 \textcolor{gray}{$\pm$ 12.4}     & 0.7 \textcolor{gray}{$\pm$ 0.9}               & 6.7 \textcolor{gray}{$\pm$ 3.4}                 & 22.7 \textcolor{gray}{$\pm$ 5.2}  &  14.7 \textcolor{gray}{$\pm$ 5.0}\\ \midrule
Train     & Mug (B, C, D)                & 61.8 \textcolor{gray}{$\pm$ 2.5}                  & 45.8 \textcolor{gray}{$\pm$ 6.7}             & \fhl{75.3}\textcolor{gray}{\fhl{ $\pm$ 2.4}         } & 48.2 \textcolor{gray}{$\pm$ 5.5}          & 72.0 \textcolor{gray}{$\pm$ 1.4}                             & 53.6 \textcolor{gray}{$\pm$ 5.4}                  & 46.7 \textcolor{gray}{$\pm$ 4.9}      & 33.1 \textcolor{gray}{$\pm$ 6.0}              & 50.7 \textcolor{gray}{$\pm$ 3.4}                & 54.9 \textcolor{gray}{$\pm$ 1.4}  &  49.3 \textcolor{gray}{$\pm$ 2.4}\\
Test      & Mug A                                   & \fhl{44.0}\textcolor{gray}{\fhl{ $\pm$ 7.5}         } & 8.7 \textcolor{gray}{$\pm$ 6.2}              & 32.0 \textcolor{gray}{$\pm$ 4.9}                  & 0.7 \textcolor{gray}{$\pm$ 0.9}           & 6.7 \textcolor{gray}{$\pm$ 3.8}                              & 14.0 \textcolor{gray}{$\pm$ 4.3}       & 14.7 \textcolor{gray}{$\pm$ 3.4}      & 9.3 \textcolor{gray}{$\pm$ 6.8}               & 0.0 \textcolor{gray}{$\pm$ 0.0}                 & 20.7 \textcolor{gray}{$\pm$ 6.2}  &  27.3 \textcolor{gray}{$\pm$ 3.4}\\ \midrule
Train     & Mug (A, C, D)                  & 61.8 \textcolor{gray}{$\pm$ 1.4}                  & 48.4 \textcolor{gray}{$\pm$ 4.9}             & 69.6 \textcolor{gray}{$\pm$ 1.4}                  & 48.0 \textcolor{gray}{$\pm$ 4.1}          & \fhl{70.4}\textcolor{gray}{\fhl{ $\pm$ 4.1}                    } & 48.4 \textcolor{gray}{$\pm$ 5.5}                & 42.2 \textcolor{gray}{$\pm$ 9.3}      & 43.3 \textcolor{gray}{$\pm$ 6.6}              & 45.8 \textcolor{gray}{$\pm$ 3.5}                & 52.9 \textcolor{gray}{$\pm$ 0.8}  &  51.6 \textcolor{gray}{$\pm$ 3.0}\\
Test      & Mug B                                 & \fhl{28.7}\textcolor{gray}{\fhl{ $\pm$ 5.2}         } & 16.7 \textcolor{gray}{$\pm$ 2.5}             & \fhl{28.7}\textcolor{gray}{\fhl{ $\pm$ 5.0}         } & 12.0 \textcolor{gray}{$\pm$ 4.9}          & 22.0 \textcolor{gray}{$\pm$ 9.1}                             & 23.3 \textcolor{gray}{$\pm$ 5.2}         & 6.7 \textcolor{gray}{$\pm$ 3.8}       & 3.3 \textcolor{gray}{$\pm$ 0.9}              & 0.0 \textcolor{gray}{$\pm$ 0.0}                 & 14.0 \textcolor{gray}{$\pm$ 1.6}  &  21.3 \textcolor{gray}{$\pm$ 5.2}\\
\midrule
\multicolumn{2}{l}{Train Avg.}        & 60.6 \textcolor{gray}{$\pm$ 2.5}                  & 44.6 \textcolor{gray}{$\pm$ 4.5}             & \fhl{71.3}\textcolor{gray}{\fhl{ $\pm$ 1.9}         } & 47.4 \textcolor{gray}{$\pm$ 2.9}          & 70.4 \textcolor{gray}{$\pm$ 3.4}                             & 49.4 \textcolor{gray}{$\pm$ 2.3}                     & 44.1 \textcolor{gray}{$\pm$ 4.4}                  & 37.4 \textcolor{gray}{$\pm$ 2.6}              & 48.0 \textcolor{gray}{$\pm$ 1.0}                & 53.9 \textcolor{gray}{$\pm$ 1.6}  &  49.9 \textcolor{gray}{$\pm$ 1.1}\\
\multicolumn{2}{l}{Out-Of-Distribution Avg.}    & \fhl{45.7}\textcolor{gray}{\fhl{ $\pm$ 1.7}         } & 14.7 \textcolor{gray}{$\pm$ 2.1}             & 29.3 \textcolor{gray}{$\pm$ 1.5}                  & 10.5 \textcolor{gray}{$\pm$ 1.1}          & 21.0 \textcolor{gray}{$\pm$ 1.1}                             & 31.2 \textcolor{gray}{$\pm$ 1.4}           & 22.0 \textcolor{gray}{$\pm$ 2.0}        & 4.8 \textcolor{gray}{$\pm$ 0.8}               & 3.2 \textcolor{gray}{$\pm$ 1.6}  & 23.7 \textcolor{gray}{$\pm$ 2.6}  &  22.3 \textcolor{gray}{$\pm$ 0.8}
\\ \bottomrule \\
\end{tabularx}
\vspace{-8pt}
\caption{\textbf{Task success rate on intra-class generalization experiment.} Results shown are the average of 3 random seeds, with 50 evaluations each, with $\pm$ standard deviation. The \fhl{highest} performing policies are highlighted.}
\label{tab:exp1}
\vspace{-12pt}
\end{table*}

%% file: tex_files/inter_table_new.tex
\begin{table*}[ht]
\scriptsize
\setlength{\tabcolsep}{1.2pt}
\centering
\begin{tabularx}{\textwidth}{llXXXXXXXXXXl}
\toprule

\multicolumn{2}{l}{Feature Extractor Method} & \multicolumn{1}{l}{DVK} & \multicolumn{2}{l}{DINO} & \multicolumn{2}{l}{SL} & \multicolumn{2}{l}{Scratch} & \multicolumn{1}{l}{MVP} & \multicolumn{1}{l}{R3M} & \multicolumn{1}{l}{VC-1} & \multicolumn{1}{l}{Soup} \\
\cmidrule(r){3-3} \cmidrule(r){4-5} \cmidrule(r){6-7} \cmidrule(r){8-9} 
\cmidrule(r){10-10} \cmidrule(r){11-11} \cmidrule{12-12} \cmidrule{13-13}
\multicolumn{2}{l}{Policy Network} & MLP & MLP & RNN & MLP & RNN & MLP & RNN & MLP & MLP & MLP & MLP \\

\midrule
\multirow{3}{*}{Train}
& Mug D & 52.0 \textcolor{gray}{$\pm$ 2.8} & 38.0 \textcolor{gray}{$\pm$ 5.7} & \shl{69.3}\textcolor{gray}{\shl{ $\pm$ 6.2}} & 37.3 \textcolor{gray}{$\pm$ 4.1}          & \fhl{70.7}\textcolor{gray}{\fhl{ $\pm$ 7.7}         } & 43.3 \textcolor{gray}{$\pm$ 6.6}& 30.7 \textcolor{gray}{$\pm$ 11.8}                                                                           &  39.3 \textcolor{gray}{$\pm$ 3.4}        & 47.3 \textcolor{gray}{$\pm$ 0.9} & 54.0 \textcolor{gray}{$\pm$ 2.8}    & 63.3 \textcolor{gray}{$\pm$ 5.0}\\
& Pan& 83.3 \textcolor{gray}{$\pm$ 2.5}& 69.3 \textcolor{gray}{$\pm$ 6.8}            & \fhl{89.3}\textcolor{gray}{\fhl{ $\pm$ 3.4}     } & 71.3 \textcolor{gray}{$\pm$ 6.2}          & \shl{88.7}\textcolor{gray}{\shl{ $\pm$ 2.5}                  } & 69.3 \textcolor{gray}{$\pm$ 3.4}& 71.3 \textcolor{gray}{$\pm$ 5.0}                                                       &  60.0 \textcolor{gray}{$\pm$ 5.9}        & 68.7 \textcolor{gray}{$\pm$ 5.0}                 & 83.3 \textcolor{gray}{$\pm$ 1.9}    & 73.3 \textcolor{gray}{$\pm$ 4.1}\\
& Screwdriver         & \fhl{91.3}\textcolor{gray}{\fhl{ $\pm$ 2.5}          } & 73.3 \textcolor{gray}{$\pm$ 5.2}            & \fhl{91.3}\textcolor{gray}{\fhl{ $\pm$ 5.2}     } & 66.7 \textcolor{gray}{$\pm$ 2.5} & \shl{89.3}\textcolor{gray}{\shl{ $\pm$ 1.9}                  } & 87.3 \textcolor{gray}{$\pm$ 3.4}           & 76.0 \textcolor{gray}{$\pm$ 7.1}                 &  59.3 \textcolor{gray}{$\pm$ 5.7}           & 79.3 \textcolor{gray}{$\pm$ 5.2}                 & 76.7 \textcolor{gray}{$\pm$ 4.1}    & 72.0 \textcolor{gray}{$\pm$ 1.6}\\
\midrule
\multirow{21}{*}{Test}
& Mug A             & \fhl{47.3}\textcolor{gray}{\fhl{ $\pm$ 0.9}          } & 10.0 \textcolor{gray}{$\pm$ 2.8}            & 3.3 \textcolor{gray}{$\pm$ 3.4}               & 6.0 \textcolor{gray}{$\pm$ 0.0}           & 11.3 \textcolor{gray}{$\pm$ 1.9}                  & 20.0 \textcolor{gray}{$\pm$ 4.3} & 8.0 \textcolor{gray}{$\pm$ 5.9}                     &  9.3 \textcolor{gray}{$\pm$ 4.1}            & 0.0 \textcolor{gray}{$\pm$ 0.0}  & \shl{22.7}\textcolor{gray}{\shl{ $\pm$ 13.1}}    & 9.3 \textcolor{gray}{$\pm$ 4.1}\\
& Mug B & \shl{30.0}\textcolor{gray}{\shl{ $\pm$ 5.7}} & 8.7 \textcolor{gray}{$\pm$ 1.9} & 21.3 \textcolor{gray}{$\pm$ 2.5} & 4.7 \textcolor{gray}{$\pm$ 2.5}           & 14.0 \textcolor{gray}{$\pm$ 1.6}                  & \fhl{32.0}\textcolor{gray}{\fhl{ $\pm$ 11.4}              } & 19.3 \textcolor{gray}{$\pm$ 3.8}                                                     &  7.3 \textcolor{gray}{$\pm$ 2.5}            & 4.0 \textcolor{gray}{$\pm$ 3.3}                  & 16.7 \textcolor{gray}{$\pm$ 6.8}    & 9.3 \textcolor{gray}{$\pm$ 4.1}\\
& Mug C           & \shl{24.7}\textcolor{gray}{\shl{ $\pm$ 3.8}                   } & 14.0 \textcolor{gray}{$\pm$ 3.3}            & 15.3 \textcolor{gray}{$\pm$ 5.2}              & 4.7 \textcolor{gray}{$\pm$ 1.9}           & 9.3 \textcolor{gray}{$\pm$ 3.4}                   & \fhl{40.7}\textcolor{gray}{\fhl{ $\pm$ 4.7}  } & 19.3 \textcolor{gray}{$\pm$ 5.7}            &  8.0 \textcolor{gray}{$\pm$ 2.8}            & 2.0 \textcolor{gray}{$\pm$ 1.6}                  & 11.3 \textcolor{gray}{$\pm$ 5.7}    & 12.0 \textcolor{gray}{$\pm$ 4.3}\\
& Teapot         & \shl{14.0}\textcolor{gray}{\shl{ $\pm$ 5.9}                   } & 4.0 \textcolor{gray}{$\pm$ 1.6}             & \fhl{15.3}\textcolor{gray}{\fhl{ $\pm$ 3.4}     } & 4.0 \textcolor{gray}{$\pm$ 3.3}           & 3.3 \textcolor{gray}{$\pm$ 0.9}                   & 5.3 \textcolor{gray}{$\pm$ 0.9}& 0.0 \textcolor{gray}{$\pm$ 0.0}                          &  0.0 \textcolor{gray}{$\pm$ 0.0}            & 0.0 \textcolor{gray}{$\pm$ 0.0}                  & 4.0 \textcolor{gray}{$\pm$ 0.0}    & 6.7 \textcolor{gray}{$\pm$ 1.9}\\
& Pitcher        & 8.0\textcolor{gray}{ $\pm$ 4.3                      } & 1.3 \textcolor{gray}{$\pm$ 0.9}             & 4.0 \textcolor{gray}{$\pm$ 4.3}               & 3.3 \textcolor{gray}{$\pm$ 0.9}           & 4.7 \textcolor{gray}{$\pm$ 2.5}                   & \shl{10.7}\textcolor{gray}{\shl{ $\pm$ 9.6} } & 7.3 \textcolor{gray}{$\pm$ 4.1}             &  5.3 \textcolor{gray}{$\pm$ 2.5}            & 6.7 \textcolor{gray}{$\pm$ 3.8}                  & \fhl{14.7}\textcolor{gray}{\fhl{ $\pm$ 0.9}}    & 0.7 \textcolor{gray}{$\pm$ 0.9}\\
& Coffee              & \shl{8.7}\textcolor{gray}{\shl{ $\pm$ 3.4}           } & \shl{8.0}\textcolor{gray}{\shl{ $\pm$ 3.3}             } & \shl{8.0}\textcolor{gray}{\shl{ $\pm$ 4.9}               } & 3.3 \textcolor{gray}{$\pm$ 1.9}           & 6.0 \textcolor{gray}{$\pm$ 3.3}                   & \shl{8.0}\textcolor{gray}{\shl{ $\pm$ 3.3}} & 4.7 \textcolor{gray}{$\pm$ 2.5}                     &  7.3 \textcolor{gray}{$\pm$ 3.8}            & 2.7 \textcolor{gray}{$\pm$ 2.5}                  & \fhl{10.0}\textcolor{gray}{\fhl{ $\pm$ 5.7}}    & 3.3 \textcolor{gray}{$\pm$ 0.9}\\
& Kidcup              & \fhl{10.0}\textcolor{gray}{\fhl{ $\pm$ 3.3}          } & 2.0 \textcolor{gray}{$\pm$ 1.6}             & 4.7 \textcolor{gray}{$\pm$ 6.6}               & 4.0 \textcolor{gray}{$\pm$ 5.7}           & 0.7 \textcolor{gray}{$\pm$ 0.9}                   & \shl{6.0}\textcolor{gray}{\shl{ $\pm$ 1.6}} & 2.7 \textcolor{gray}{$\pm$ 2.5}                     &  0.0 \textcolor{gray}{$\pm$ 0.0}            & 0.0 \textcolor{gray}{$\pm$ 0.0}                  & \shl{6.7}\textcolor{gray}{\shl{ $\pm$ 8.1}}    & 4.7 \textcolor{gray}{$\pm$ 2.5}\\
\cmidrule{3-13}
& Hammer              & 50.7 \textcolor{gray}{$\pm$ 4.1}                   & 48.7 \textcolor{gray}{$\pm$ 10.9}           & \fhl{66.7}\textcolor{gray}{\fhl{ $\pm$ 4.1}     } & 42.7 \textcolor{gray}{$\pm$ 5.7}          & \shl{62.7}\textcolor{gray}{\shl{ $\pm$ 9.3}                  } & 58.7 \textcolor{gray}{$\pm$ 4.7}& 48.7 \textcolor{gray}{$\pm$ 5.0}                    &  16.0 \textcolor{gray}{$\pm$ 2.8}            & 19.3 \textcolor{gray}{$\pm$ 5.7}                 & \shl{64.0}\textcolor{gray}{\shl{ $\pm$ 3.3}}    & 50.0 \textcolor{gray}{$\pm$ 1.6}\\
& Spatula             & 58.7 \textcolor{gray}{$\pm$ 9.3}                   & 35.3 \textcolor{gray}{$\pm$ 0.9}            & 43.3 \textcolor{gray}{$\pm$ 10.9}             & 14.7 \textcolor{gray}{$\pm$ 4.1}          & 32.7 \textcolor{gray}{$\pm$ 3.8}                  & \fhl{72.7}\textcolor{gray}{\fhl{ $\pm$ 3.4} } & \shl{62.0}\textcolor{gray}{\shl{ $\pm$ 5.9}          } &  20.0 \textcolor{gray}{$\pm$ 2.8}           & 0.0 \textcolor{gray}{$\pm$ 0.0}                  &  39.3 \textcolor{gray}{$\pm$ 4.1}    & 30.7 \textcolor{gray}{$\pm$ 1.9}\\
& Mag. Glass            & \fhl{28.7}\textcolor{gray}{\fhl{ $\pm$ 1.9}          } & 0.7 \textcolor{gray}{$\pm$ 0.9}             & 2.7 \textcolor{gray}{$\pm$ 0.9}               & 4.7 \textcolor{gray}{$\pm$ 0.9}           & 11.3 \textcolor{gray}{$\pm$ 5.2}                  & \shl{14.0}\textcolor{gray}{\shl{ $\pm$ 5.9}} & 2.0 \textcolor{gray}{$\pm$ 2.8}                   &  6.0 \textcolor{gray}{$\pm$ 4.3}            & 0.0 \textcolor{gray}{$\pm$ 0.0}                  & 2.7 \textcolor{gray}{$\pm$ 0.9}    & 2.0 \textcolor{gray}{$\pm$ 1.9}\\
& Hair Straightener    & \shl{50.0}\textcolor{gray}{\shl{ $\pm$ 4.3}                   } & 14.7 \textcolor{gray}{$\pm$ 0.9}            & 18.0 \textcolor{gray}{$\pm$ 7.5}              & 28.7 \textcolor{gray}{$\pm$ 8.4}          & 32.7 \textcolor{gray}{$\pm$ 6.8}                  & \fhl{55.3}\textcolor{gray}{\fhl{ $\pm$ 9.4}} & 37.3 \textcolor{gray}{$\pm$ 5.2}          &  18.0 \textcolor{gray}{$\pm$ 2.8}           & 8.0 \textcolor{gray}{$\pm$ 3.3}                 & \shl{51.3}\textcolor{gray}{\shl{ $\pm$ 7.4}}    & 36.7 \textcolor{gray}{$\pm$ 3.8}\\
& Spoonrest           & \shl{51.3}\textcolor{gray}{\shl{ $\pm$ 10.9}                  } & 38.0 \textcolor{gray}{$\pm$ 5.9}            & \fhl{56.7}\textcolor{gray}{\fhl{ $\pm$ 5.0}     } & 27.3 \textcolor{gray}{$\pm$ 3.4}          & 43.3 \textcolor{gray}{$\pm$ 5.2}                  & 38.0 \textcolor{gray}{$\pm$ 4.3}& 10.7 \textcolor{gray}{$\pm$ 6.2}                    &  4.7 \textcolor{gray}{$\pm$ 2.5}            & 0.0 \textcolor{gray}{$\pm$ 0.0}                  & 28.0 \textcolor{gray}{$\pm$ 3.3}    & 12.0 \textcolor{gray}{$\pm$ 4.9}\\
& DustPan             & \fhl{32.0}\textcolor{gray}{\fhl{ $\pm$ 5.7}          } & 25.3 \textcolor{gray}{$\pm$ 1.9}            & \shl{29.3}\textcolor{gray}{\shl{ $\pm$ 2.5}              } & 9.3 \textcolor{gray}{$\pm$ 5.0}           & 22.7 \textcolor{gray}{$\pm$ 3.4}                  & 18.7 \textcolor{gray}{$\pm$ 7.7}& 8.0 \textcolor{gray}{$\pm$ 3.3}                     &  2.0 \textcolor{gray}{$\pm$ 2.8}           & 0.0 \textcolor{gray}{$\pm$ 0.0}                 & 8.0 \textcolor{gray}{$\pm$ 3.3}    & 16.7 \textcolor{gray}{$\pm$ 6.8}\\
                                                                               \cmidrule{3-13}
& Basket              & \shl{16.0}\textcolor{gray}{\shl{ $\pm$ 2.8}                   } & 12.7 \textcolor{gray}{$\pm$ 8.4}            & 4.0 \textcolor{gray}{$\pm$ 5.7}               & 12.0 \textcolor{gray}{$\pm$ 7.5}          & \fhl{18.7}\textcolor{gray}{\fhl{ $\pm$ 6.2}         } & \shl{16.0}\textcolor{gray}{\shl{ $\pm$ 5.9}} & 6.0 \textcolor{gray}{$\pm$ 8.5}                     &  0.7 \textcolor{gray}{$\pm$ 0.9}            & 0.7 \textcolor{gray}{$\pm$ 0.9}                 & 14.7 \textcolor{gray}{$\pm$ 6.2}    & 8.7 \textcolor{gray}{$\pm$ 6.6}\\
& Bell                & \fhl{30.0}\textcolor{gray}{\fhl{ $\pm$ 10.2}         } & 15.3 \textcolor{gray}{$\pm$ 4.7}            & 12.7 \textcolor{gray}{$\pm$ 9.0}              & 6.7 \textcolor{gray}{$\pm$ 0.9}           & 22.0 \textcolor{gray}{$\pm$ 8.6}                  & 22.0 \textcolor{gray}{$\pm$ 13.4}  & \shl{28.7}\textcolor{gray}{\shl{ $\pm$ 14.8}                } &  20.7 \textcolor{gray}{$\pm$ 3.8}            & 1.3 \textcolor{gray}{$\pm$ 0.9}                & 8.7 \textcolor{gray}{$\pm$ 5.7}    & 8.7 \textcolor{gray}{$\pm$ 12.3}\\
&C-Clamp              & 5.3 \textcolor{gray}{$\pm$ 4.1}                    & 10.0 \textcolor{gray}{$\pm$ 5.9}            & 14.0 \textcolor{gray}{$\pm$ 4.3}              & 2.7 \textcolor{gray}{$\pm$ 3.8}           & \fhl{21.3}\textcolor{gray}{\fhl{ $\pm$ 9.0}         } & \shl{20.7}\textcolor{gray}{\shl{ $\pm$ 2.5}} & 6.7 \textcolor{gray}{$\pm$ 5.0}                    &  7.3 \textcolor{gray}{$\pm$ 2.5}            & 3.3 \textcolor{gray}{$\pm$ 0.9}                 & 10.0 \textcolor{gray}{$\pm$ 3.3}    & 7.3 \textcolor{gray}{$\pm$ 7.5}\\
& Honeydip            & 2.7 \textcolor{gray}{$\pm$ 2.5}                    & 16.7 \textcolor{gray}{$\pm$ 10.5}           & \fhl{26.7}\textcolor{gray}{\fhl{ $\pm$ 16.4}    } & \shl{16.7}\textcolor{gray}{\shl{ $\pm$ 6.2}          } & 6.7 \textcolor{gray}{$\pm$ 5.2}                   & 2.7 \textcolor{gray}{$\pm$ 1.9}& 4.0 \textcolor{gray}{$\pm$ 5.7}                      &  0.7 \textcolor{gray}{$\pm$ 0.9}            & 0.7 \textcolor{gray}{$\pm$ 0.9}                 & 7.3 \textcolor{gray}{$\pm$ 4.1}    & 0.7 \textcolor{gray}{$\pm$ 0.9}\\
& Shoe1               & \fhl{30.0}\textcolor{gray}{\fhl{ $\pm$ 2.8}          } & 24.7 \textcolor{gray}{$\pm$ 9.6}            & \shl{27.3}\textcolor{gray}{\shl{ $\pm$ 6.8}              } & 20.0 \textcolor{gray}{$\pm$ 3.3}          & 21.3 \textcolor{gray}{$\pm$ 6.6}                  & 24.0 \textcolor{gray}{$\pm$ 5.9}& 10.0 \textcolor{gray}{$\pm$ 4.9}                    &  9.3 \textcolor{gray}{$\pm$ 4.7}            & 1.3 \textcolor{gray}{$\pm$ 0.9}                 & 14.0 \textcolor{gray}{$\pm$ 6.5}    & 12.0 \textcolor{gray}{$\pm$ 8.2}\\
& Shoe2               & \shl{28.0}\textcolor{gray}{\shl{ $\pm$ 5.7}                   } & 22.0 \textcolor{gray}{$\pm$ 1.6}            & \fhl{42.0}\textcolor{gray}{\fhl{ $\pm$ 7.1}     } & 13.3 \textcolor{gray}{$\pm$ 4.1}          & 26.0 \textcolor{gray}{$\pm$ 5.9}                  & 17.3 \textcolor{gray}{$\pm$ 2.5}& 10.0 \textcolor{gray}{$\pm$ 3.3}                    &  11.3 \textcolor{gray}{$\pm$ 4.1}            & 0.0 \textcolor{gray}{$\pm$ 0.0}                 & \shl{28.7}\textcolor{gray}{\shl{ $\pm$ 3.4}}     & 26.0 \textcolor{gray}{$\pm$ 5.9}\\
& Rattle              & \fhl{45.3}\textcolor{gray}{\fhl{ $\pm$ 13.9}         } & 12.7 \textcolor{gray}{$\pm$ 1.9}            & 13.3 \textcolor{gray}{$\pm$ 6.6}              & 14.0 \textcolor{gray}{$\pm$ 8.5}          & 16.0 \textcolor{gray}{$\pm$ 4.9}                  & \shl{24.0}\textcolor{gray}{\shl{ $\pm$ 8.2}} & 13.3 \textcolor{gray}{$\pm$ 10.5}                   &  2.7 \textcolor{gray}{$\pm$ 0.9}           & 8.0 \textcolor{gray}{$\pm$ 2.8}                 & 11.3 \textcolor{gray}{$\pm$ 3.8}    & 11.3 \textcolor{gray}{$\pm$ 5.0}\\
& Lion Toy             & \fhl{20.7}\textcolor{gray}{\fhl{ $\pm$ 4.7}          } & 2.7 \textcolor{gray}{$\pm$ 2.5}             & 4.0 \textcolor{gray}{$\pm$ 4.3}               & 18.7\textcolor{gray}{ $\pm$ 7.7          } & 9.3 \textcolor{gray}{$\pm$ 0.9}                   & 16.0 \textcolor{gray}{$\pm$ 4.3}& 4.0 \textcolor{gray}{$\pm$ 1.6}                    &  4.0 \textcolor{gray}{$\pm$ 2.8}             & 0.7 \textcolor{gray}{$\pm$ 0.9}                 & \shl{19.3}\textcolor{gray}{\shl{$ \pm$ 4.1}}    & 7.3 \textcolor{gray}{$\pm$ 1.9}\\
\midrule
\multicolumn{2}{l}{Train Avg.}     & 75.6 \textcolor{gray}{$\pm$ 1.7}                   & 60.2 \textcolor{gray}{$\pm$ 5.5}            & \fhl{83.3}\textcolor{gray}{\fhl{ $\pm$ 3.4}     } & 58.4 \textcolor{gray}{$\pm$ 2.1}          & \shl{82.9}\textcolor{gray}{\shl{ $\pm$ 2.1}                  } & 66.7 \textcolor{gray}{$\pm$ 2.2}& 59.3 \textcolor{gray}{$\pm$ 4.3} & 52.9 \textcolor{gray}{$\pm$ 1.7}             & 65.1 \textcolor{gray}{$\pm$ 1.7}                & 71.3 \textcolor{gray}{$\pm$ 0.5}     & 69.6 \textcolor{gray}{$\pm$ 2.7}\\
\multicolumn{2}{l}{Out-Of-Distribution Avg.} & \fhl{28.2}\textcolor{gray}{\fhl{ $\pm$ 1.9}          } & 15.6 \textcolor{gray}{$\pm$ 0.5}            & \shl{20.6}\textcolor{gray}{\shl{ $\pm$ 3.7} } & 12.4 \textcolor{gray}{$\pm$ 1.0}          & 18.9 \textcolor{gray}{$\pm$ 1.2}                  & 24.9 \textcolor{gray}{$\pm$ 0.7}& 14.9 \textcolor{gray}{$\pm$ 1.6}     & 7.7 \textcolor{gray}{$\pm$ 1.2}            & 2.8 \textcolor{gray}{$\pm$ 0.4}  & 18.7 \textcolor{gray}{$\pm$ 2.4}     & 13.3 \textcolor{gray}{$\pm$ 2.2}\\
\bottomrule
\end{tabularx}
\caption{\textbf{Task success rate on the inter-class generalization experiment.} Results shown are the average of 3 random seeds, with 50 evaluations each, with $\pm$ standard deviation. The \fhl{highest} and \shl{second-highest} performing policies are highlighted.}
\label{tab:exp3}
\vspace{-12pt}
\end{table*}

%% file: tex_files/ablation_table.tex
\begin{table}[t]
\centering
\setlength{\tabcolsep}{2pt}
\newcommand{\pp}{\phantom{-}}
\scriptsize
\begin{tabularx}{0.45\textwidth}{l*4{>{\centering\arraybackslash}X}}
\toprule
& \multicolumn{2}{c}{Intra-class} & \multicolumn{2}{c}{Inter-class}\\
\cmidrule(r){2-3}  \cmidrule(r){4-5}
& Train & Test & Train & Test  \\
\midrule
\multicolumn{5}{l}{\quad Feature embedding (Default: DVK)} \\ 
\midrule
DINO ViT$_\texttt{[CLS]}$      & -10.8 & -15.0   & -26.0 & -13.9 \\
DINO ViT$_{\texttt{[CLS]} + \text{GeMPool}}$      & -14.3 &  -17.4     & -32.5 & -17.1 \\
\midrule
\multicolumn{5}{l}{\quad Candidate clusters/ref. features M/m (Default: $M=100, m=50$)} \\ 
\midrule
$M=25, m=12$     & -11.0          &     -13.0  & \po-3.6 & \po-2.2\\
$M=50, m=25$      &    \po-5.2      &     \po-2.9   & \po-2.0 & +1.6\\
$M=200, m=100$      &    +1.6     &     \po-0.2    & +1.3 & \po-0.9\\
$M=400, m=200$      &     +3.7    &    +3.0   & \po-0.3 & +0.1\\
\bottomrule
\end{tabularx}
 \caption{\textbf{Average success rates for different ablations of DVK in terms of \emph{percent difference}}.}
 \label{table:ablat}
\vspace{-16pt}
\end{table}